\documentclass{article}

\usepackage{neurips_2019}

\usepackage[utf8]{inputenc} % allow utf-8 input
\usepackage[T1]{fontenc}    % use 8-bit T1 fonts
\usepackage{hyperref}       % hyperlinks
\usepackage{url}            % simple URL typesetting
\usepackage{graphicx}
\usepackage{subfigure}
\usepackage{booktabs}       % professional-quality tables
\usepackage{amsfonts}       % blackboard math symbols
\usepackage{nicefrac}       % compact symbols for 1/2, etc.
\usepackage[ruled,vlined]{algorithm2e}
\usepackage{caption}
\usepackage{lineno}
\usepackage{microtype}      % microtypography
\nolinenumbers
\title{Self-semi-supervised Learning to Learn from Noisy Labeled Data}

\begin{document}

\maketitle

\section{Introduction}
The remarkable success of today’s deep neural networks (DNN) highly depends on a massive number of correctly labeled data. However, it is rather costly to obtain high-quality human-labeled data, leading to the active research area of training models robust to noisy labels. To achieve this goal, on the one hand, many papers have been dedicated to differentiating noisy labels from clean ones to increase the generalization of DNN. On the other hand, the increasingly prevalent methods of self-semi-supervised learning have been proven to benefit the tasks when labels are incomplete. By “semi” we regard the wrongly labeled data detected as unlabelled data; by “self” we choose a self-supervised technique to conduct semi-supervised learning. In this project, we designed methods to more accurately differentiate clean and noisy labels and borrowed the wisdom of self-semi-supervised learning to train noisy-labeled data.

\section{State of Art}
Existing studies on deep learning with noisy-labeled data sets focus on either proposing robust loss correction methods or modifying training architectures to fully utilize noisy data. Most of the studies attempt to re-weight the contribution of noisy-labeled data, requiring a well-designed criteria to identify noisy labels during training. Recent studies suggest salient signal of different fitting pattern with clean and noisy labels; for instance, noisy labels take longer to learn than clean labels, so theoretically it is possible to distinguish the clean and noisy samples from loss distributions (Zhang et al. 2016 [1]). In Arazo et al. 2019 [2], a two component Beta/Gaussian mixture model is fit to loss distribution and data with lower loss is treated as clean. Geoff et al. 2020 [3] proposes a statistical method to measure the confidence of prediction to obtain a robust indicator of mislabeled samples. Our work further explored based on these insights, in which we hypothesize that salient signal beyond loss difference can be obtained through training. 

Ideally, if noisy labels and clean labels are accurately separated, then the issue with noisy labeled data sets can be transformed to semi-supervised learning. One popular framework (Arazo et al. 2019 [2], Li et al. 2020 [4]) is to replace labels detected as noisy with predicted labels by well-designed "guess and refinement". Another popular framework (e.g., S4L framework in Zhai et al. 2019 [4]) is to minimize the total loss of predicting the label of the labeled data and predicting the self-supervised class of the unlabelled data (or of the full data). 

The combination of the two studied area is under-explored. While Zhai et al. 2019 [4] applies self-semi-supervision in the scenario where each sample is deterministic to be either labeled or unlabelled, we attempt to transfer this technique to our scenario where whether each sample is labeled correctly or wrongly is unobserved but detectable.

\section{Methodology}
The data set we used is CIFAR-10. We split the CIFAR-10's original training set to a training set consisting of 45000 images and a validation set consisting of 5000 images. To simulate a data set with noisy labels, we randomly sampled $r$-percent images from the training set, and randomly selected integers from 0 to 9 as their labels. Below we explained our two contributions in detail.

\subsection{Loss-and-Confidence Noisy Detection}
As discussed in the previous work (Arazo et al. 2019 [2]), DNN always fit clean label first, so data with clean label in general has lower loss than that with noisy label in an early stage. Hence, the author fits a two component Gaussian Mixture Model (GMM) on the loss distribution which can output samples' probabilities of being clean and noisy. Unfortunately, there are two issues with this method. First, GMM is not good at classifying labels whose probability of being clean is close to that of being noisy. Second, DNN starts to memorize noisy labels before it learns enough information from clean labels, making the classification of labels inaccurate. Therefore, in this project, we innovate on noisy-label detection methods to deal with these two issues.

Our first innovation is the incorporation of a GMM fitted on the confidence of prediction (confirmation bias), along with one fitted on the cross-entropy loss as before. Based on our experiments, clean-labeled data shows higher confidence of its prediction than noisy-labeled data, and this difference in confidence remains throughout the training. Therefore, we can derive more knowledge on whether a label is clean or noisy though this new GMM.

Our second innovation is the ways to combine information from the two GMMs above. We have tried two different decision strategies, hard and elastic decision, as follows:

\subsubsection{Hard Decision}
The simplest and most straightforward approach is to strictly treat a sample as clean only when both GMMs predict it as clean, as listed below:
$$\mathbb{I}_{clean}[\min(GMM_{loss},GMM_{confidence})>0.5]$$
As shown in the Figure 1.a, the performance of this decision strategy improves noisy label detection ability compared that of $GMM_{loss}>0.5$. The clean label prediction accuracy is higher and the problem of over-fitting to noisy label is alleviated. This improvement indicates a more accurate separation of clean and noisy labels especially in late epoch. However, the hard decision will only regard very few of the data as clean, preventing the DNN from gaining sufficient information from clean labels in early stage. To further improve the performance, we need a decision strategy that is less strict at the beginning and as strict when the DNN tend to memorize noisy information. 

\subsubsection{Elastic Decision}
To overcome the shortcoming of the hard decision strategy, we investigated using LogSumExp (LSE) to smoothly approximate the  minimum/maximum of two value defined below:
$$LSE(\alpha,i,j) = -\alpha \log(\sum_{d}-\frac{1}{2}(\exp(\frac{i_d}{\alpha})+\exp(\frac{j_d}{\alpha}))) $$
Where $\alpha$ is a parameter (a.k.a, temperature) to approximate the maximum when $\alpha:0_- \rightarrow -\infty$ or minimum when $\alpha:0_+ \rightarrow +\infty$. We can set a loose decision at the beginning of our training phase to allow DNN to learn sufficiently. In later epochs, however, to prevent DNN from fitting noisy-labeled data, we would need strict hard decision. Inspired from Loshchilov et al. 2016 [6], we applied a cosine ramp-down function to tune the LSE temperature $\alpha$ along with training. In which, with the increasing number of epoch, our decision elastically dropped from maximum of the two GMM prediction (loosest) to decision to minimum of the two (strictest). The cosine ramping down LSE value along the epochs is shown in Figure 1.b.

\subsection{Self-Semi-Supervised Learning}
After detecting noisy labels, we innovated on the models' objective functions so that our models can accommodate data with various probabilities to be clean. Inspired by training on partially-labeled data with self-semi-supervised learning, we proposed an approach to apply self-supervised technique on noisy labels, and in this case we adopt rotation recognition as the self-supervised learning task. Before fed into the model, each image $x$ is randomly rotated an angle from $\lbrace 0^{\circ}, 90^{\circ}, 180^{\circ}, 270^{\circ} \rbrace$, and tagged with the label of rotation $y_{rot}$ besides the label of classification $y_{class}$. In our architecture, two parallel linear layers at the end of Pre-Activated ResNet18 are used for two different tasks: predicting classification label $y_{class}$ and predicting rotation label $y_{rot}$. 

Overall, we aimed at preventing the model from memorizing the wrong information of noisy labels by distracting the samples detected to be noisy with the self-supervised learning task. There are two different methods to realize the distraction: regularizing the model through self-supervised loss, or separating the training set to train clean and noisy samples with the classification task and the rotation label prediction task respectively.

\subsubsection{Probability-based Regularization Model}
Loss function of the regularization model has the form of:
$$L_{reg}(y, \hat y, \theta_{reg}) = CE(y_{class}, \hat y_{class}) + \alpha (1-\lambda) CE(y_{rot}, \hat y_{rot})$$
where $y = \lbrace y_{class}, y_{rot} \rbrace, \hat y = \lbrace \hat y_{class}, \hat y_{rot} \rbrace, \theta_{reg} = \lbrace \alpha, \lambda \rbrace$. $\lambda$ is the probability that $y$ is clean predicted via noisy label detection, and $\alpha$ controlling the weight of regularization form can be tuned. As a result, samples with higher probability to be noisy are more targeted towards the self-supervised learning task.

\subsubsection{Noisy-or-clean Separation Model}
Loss function of the separation model has the form of:
$$L_{sep}(y, \hat y, \theta_{sep}) = CE(y_{class}, \hat y_{class})\mathbb{I}(\lambda \ge c) + CE(y_{rot}, \hat y_{rot})\mathbb{I}(\lambda < c)$$
where $y = \lbrace y_{class}, y_{rot} \rbrace, \hat y = \lbrace \hat y_{class}, \hat y_{rot} \rbrace, \theta_{sep} = \lbrace c, \lambda \rbrace$. $c$ defines the cutoff point at which we split the training samples to the clean-labeled and the noisy-labeled based on their probabilities resulted from noisy label detection. Compared to the regularization model, this model completely discards the noisy-labeled samples from training the image classification task. Therefore, the success of this method is highly dependent on the correct prediction of whether a label is clean or noisy, as demonstrated by the improved performance as we detected noisy labels using elastic decision strategy.

\begin{algorithm}
\SetAlgoLined
    \SetKwInOut{Input}{Input}
    \SetKwInOut{Setting}{Setting}
    \Input{Training set $\mathcal{X}, y_{class}$}
    \Setting{learning rate, criteria $c$, regularization weight $\alpha$, probability for clean data $\lambda = 1$, warm-up epochs, maximum epochs}
    \nl \If{warm up}{
     \nl \For{epoch < warmup epoch}{
       \nl $\hat y_{class}, \hat y_{rot}$ $\leftarrow$ Neural Network with $\mathcal{X}$ as input\;
       \nl SGD with $L_{baseline}(y_{class}, \hat y_{class}) = CE(y_{class}, \hat y_{class})$ \;
}
\nl $\lambda \leftarrow$ Noisy Label Detection Model($\mathcal{X}, y_{class}$).
}
     \nl \For{epoch < maximum epoch}{
       \nl randomly rotate each $x^{i}$ with one of $\lbrace 0^{\circ}, 90^{\circ}, 180^{\circ}, 270^{\circ} \rbrace$ to form $\mathcal{X}_{rot}$ and record $y_{rot}$\;
       \nl $\hat y_{class}, \hat y_{rot}$ $\leftarrow$ Neural Network with $\mathcal{X}_{rot}$ as input\;
       \nl SGD with $L_{reg}(y, \hat y, \theta_{reg})$, $\theta_{reg} = \lbrace \alpha, \lambda \rbrace$,
       or $L_{sep}(y, \hat y, \theta_{sep})$, $\theta_{sep} = \lbrace c, \lambda \rbrace$ \;
\nl $\lambda \leftarrow$ Noisy Label Detection Model($\mathcal{X}, y_{class}$).
}
    \caption{A General Framework for Regularization Model or Separation Model}

\end{algorithm}

\section{Results}
\subsection{Loss and Confidence Bump Decision Noisy Detection}
We explored the noisy detection capacities of our proposed decision. At each epoch, the weights computed by our decision policy will make a prediction on differentiating clean from mislabeled data. The accuracy of clean label prediction is compared among different proposed decision, shown in Figure 1.b. Hard decision outperformed among elastic decision and previous loss-only decision which indicates the salient signal from prediction confidence change improve the noisy detection capacity. 
As the lack of clean data problem discussed in section 3.1.1, we evaluated the percentage of predicted clean data which is around $31.5\%$ for hard decision (under 0.4 noise rate). From the test accuracy comparison (shown in Figure 1.c), we verified our hypothesis, even though hard decision performed better than elastic decision in noisy detection, elastic decision is better at promoting DNN to learn enough information from clean labels while maintaining a decent noisy label detection capacity.

\begin{figure}
    \centering
    \subfigure[]{\includegraphics[width=0.4\textwidth]{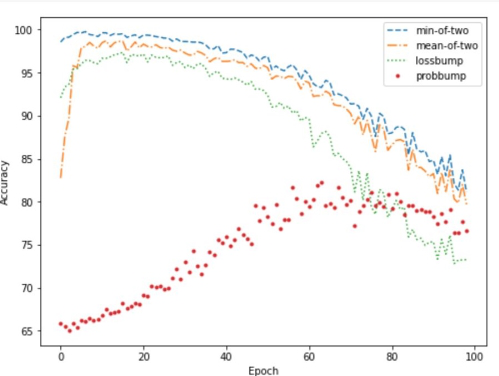}} 
    \subfigure[]{\includegraphics[width=0.4\textwidth]{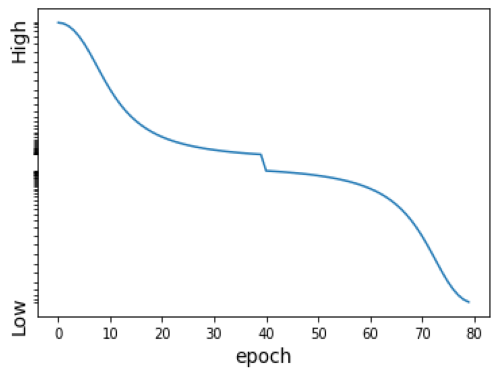}} 
    \subfigure[]{\includegraphics[width=0.4\textwidth]{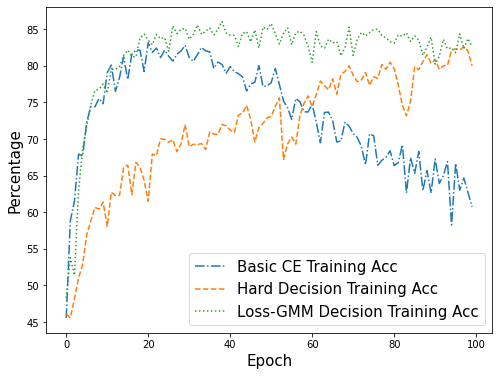}}
    \caption{(a) Clean Label prediction accuracy (b) Cosine Ramping down LSE can approximate maximum at start and minimum at the end (c) Test Accuracy trained only with predicted clean data}
\end{figure}

\subsection{Self-Semi-Supervised Learning}
Due to the augmentation (rotation) of data, there are two approaches to predict classification label for each image. We can either generate one prediction for each image without any rotation (1-image prediction) or average four predictions for each image rotated four times (4-rotation prediction). We conduct four controlled experiments and Figure 2.a shows the learning curves of several models with tuned hyper parameters on the validation set:

 \textbf{Baseline vs. Augmentation:} to test whether the data augmentation by rotation improves over the baseline model, we trained an augmentation model using the same loss function as the baseline model, i.e., $L(y_{class}, \hat y_{class}) = CE(y_{class}, \hat y_{class})$. The best validation accuracy the augmentation model achieved is roughly the same as the baseline model for 1-image prediction, but is around 2-percent higher for 4-rotation prediction.  
 
 \textbf{Augmentation vs. Regularization:} to test whether incorporating the representation learning (rotation) distracts the model from learning noisy labels, we trained the regularization model which performs obviously better then the augmentation model. The validation accuracy of the best regularization model (after parameter tuning) is increased by about 3 percents. This observation is consistent with our hypothesis that adding a regularization term weighted by the samples' probabilities of being clean can distract the model from memorizing noisy-labelled samples in an early stage. 
 
 \textbf{Regularization vs. Regularization with warm-up:} warm-up in the first 20 epoch further enhances the regularization model, and this model achieved the best validation accuracy of $71.18\%$ with 4-rotation prediction and $69.34\%$ with 1-image prediction. It makes sense intuitively because in the warm-up process, the model fits the data with clean labels rapidly without any potential distractions from the rotation recognition task, and therefore generates a plausible loss distribution to form the initial probability $\lambda$ at epoch 21 when we incorporate the representation learning.
 
 \textbf{Augmentation vs. Separation:} besides the augmentation model, we also train a separation model. This model is more conservative than the regularization model in the sense that the model will learn nothing from the labels predicted as noisy. Consequently, it fits less incorrect information, so its learning curve on the validation set shows no decline in at least 300 epochs. However, the validation accuracy of the separation model is lower than that of the regularization model (though still higher than that of baseline model), due to the loss of correct information when the separation model recognizes clean labels as noisy. 
\begin{figure}
    \centering
    \subfigure[]{\includegraphics[width=0.5\textwidth, height=5cm ]{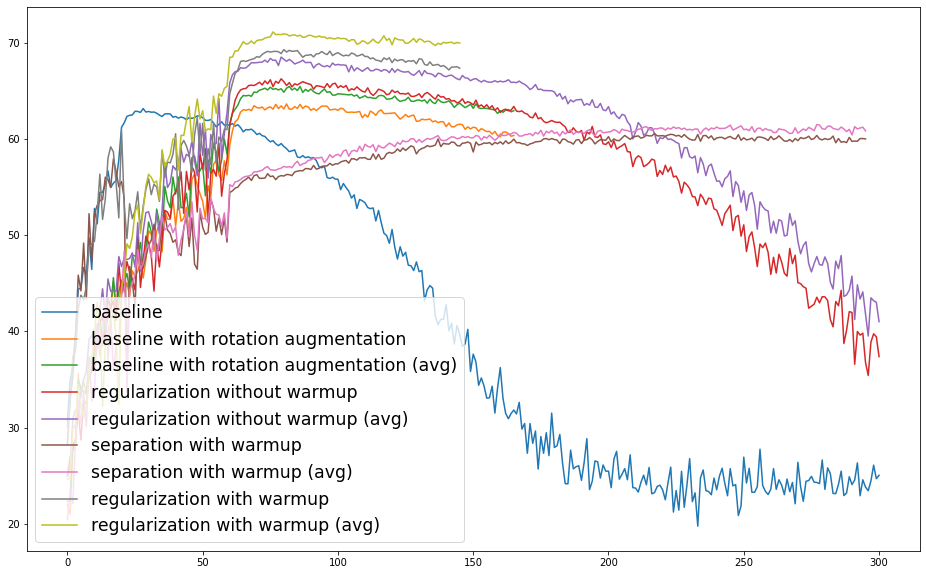}} 
    \subfigure[]{\includegraphics[width=0.4\textwidth, height=5cm]{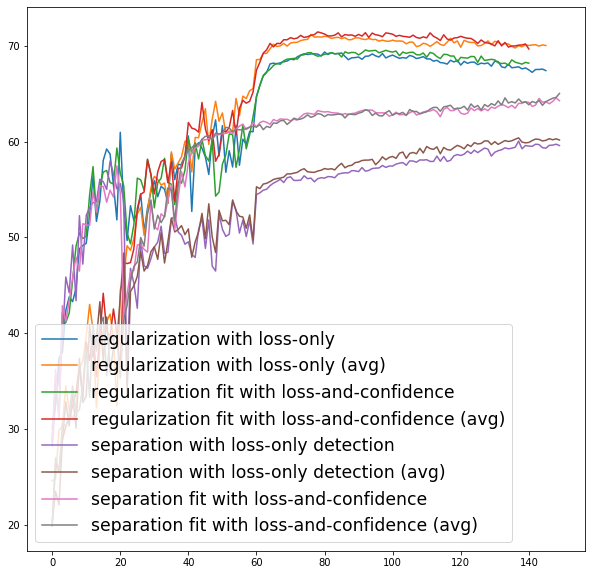}} 
    \caption{(a) validation accuracy over models using noisy-detection with only CE loss. Regularization models dominate the baseline model, while Separation models show little memorization to noisy labels at the first 300 epochs. (b) validation accuracy comparison between models using noisy-detection with 2 different methods. The new method has a lot impact on the Separation model. Based on previous experiment results, we run only 150 epoch.}
\end{figure}

\subsection{Combination of Two Approaches}
We attempted to combine the two approaches in 3.1 and 3.2 to achieve a better accuracy resulted from the more accurate noisy-labels detection and the self-supervised learning technique. As Figure 2.b shows, using the Loss-and-Confidence Noisy Detection with Elastic Decision instead of Loss-only Noisy Detection we used in the subsection 4.2, the validation accuracy of both regularization model and separation model has increased. The improvement is especially significant for the separation model more sensitive to the quality of noisy labels detection. Table 1 shows the evaluation of baseline model, regularization model and separation model using different noisy labels detection methods, and finally the best test accuracy is improved to over $69\%$ from $63.04\%$ (the baseline model).

\begin{table}
  \caption{Results for Self-semi-supervised Learning (Noise rate = 0.8)}
  \label{sample-table}
  \centering
  \resizebox{\columnwidth}{!}{\begin{tabular}{lllll}
    \toprule
    \multicolumn{1}{r}{} & \multicolumn{2}{c}{Validation Accuracy} & \multicolumn{2}{c}{Test Accuracy}                   \\
    \cmidrule(r){2-5}
    Model Name     & 1-img prediction    & 4-rot prediction     & 1-img prediction     & 4-rot prediction \\
    \midrule
    Baseline   & 63.20     & \     & 63.04     & \    \\
    Reg with loss      & 69.34      & 71.18     & 67.61     & \textbf{69.49}    \\
    Reg with loss and conf     & \textbf{69.56}     & \textbf{71.44}     & \textbf{67.93}     & 69.32    \\
    Sep with loss     & 60.04     & 60.38     & 57.92     & 58.32    \\
    Sep with loss and conf     & 64.56     & 65.02     & 62.59     & 63.12    \\
    \bottomrule
  \end{tabular}}
\end{table}

\section{Discussion}
Via application of self-semi-supervised learning on the context of data set with noisy labels, we achieved more accurate classification results. It deserves to discuss in which mechanism the improvement is sourced from.

First, observed from the loss distribution variation, the CE-loss drop dramatically and the overlapping between noisy and clean bumps is trivial at the beginning of training. Thus, we can loose our noisy label detection policy to have more data trained with the image classification task. Using a gradually decreasing function to constrain our noisy label decision will avoid both overshooting to lack of data problem and noisy label memorization.

Second, introduction of rotation recognition task strengthens our model in multiple ways. The augmentation of data set generated by rotation allows prediction made by averaging on output of 4 rotated images. Training simultaneously on the classification task and the rotation recognition task distract the model from fitting samples with noisy labels by making these sample focus more on the latter task. Another potential benefit is that through the self-supervised learning task, perhaps Deep Neural Network can learn about useful visual representation of image, since self-supervision on completely unlabeled data has been proved to be effective on a broad range of tasks. Besides, it presents two different patterns on the learning curves of Regularization models and Separation models, which may caused by different ways of handling information. While Regularization models keep using all information in training, Separation models discard labels detected to be noisy, including correct and incorrect information. Our hypothesis about the functional mechanisms underlying these results deserve further exploration with theoretical supports and experiments.

Our models make a complementary to prevailing models for addressing noisy labels issue, which usually modify noisy labels detected with predicted labels in training process, and thus increase correct information in training set. Not using predicted labels avoids the possibility to change labels from clean to noisy, but at the same time abandon the chance to acquire more correct information. We expect to combine these techniques in processing noisy labels with our models in future works.

\section*{References}

\small
[1] Zhang, Chiyuan, et al. "Understanding deep learning requires rethinking generalization." arXiv preprint arXiv:1611.03530 (2016).

[2] Arazo, Eric, et al. "Unsupervised label noise modeling and loss correction." arXiv preprint arXiv:1904.11238 (2019).

[3] Pleiss, Geoff, et al. "Identifying Mislabeled Data using the Area Under the Margin Ranking." arXiv preprint arXiv:2001.10528 (2020).

[4] Zhai, Xiaohua, et al. "$S^\mathbf {4} L$: Self-Supervised Semi-Supervised Learning." arXiv preprint arXiv:1905.03670 (2019).

[5] Li, Junnan, Richard Socher, and Steven CH Hoi. "Dividemix: Learning with noisy labels as semi-supervised learning." arXiv preprint arXiv:2002.07394 (2020).

[6] Loshchilov, Ilya, and Frank Hutter. "Sgdr: Stochastic gradient descent with warm restarts." arXiv preprint arXiv:1608.03983 (2016).
\end{document}